%% file: conference_101719.tex
\documentclass[conference]{ieeeconf}
\IEEEoverridecommandlockouts
\usepackage{cite}
\usepackage{amsmath,amssymb,amsfonts}
\usepackage{algorithmic}
\usepackage{graphicx}
\usepackage{textcomp}
\usepackage{xcolor}
\usepackage{soul}
\usepackage{multirow} 
\def\BibTeX{{\rm B\kern-.05em{\sc i\kern-.025em b}\kern-.08em
    T\kern-.1667em\lower.7ex\hbox{E}\kern-.125emX}}
\begin{document}

\title{\LARGE \bf Knowledge Distillation for Feature Extraction in Underwater VSLAM

\thanks{$^{1}$The authors are with the Department of Electrical and Electronic Engineering, The University of Melbourne.
$^{2}$The authors are with the School of Mathematics and Statistics, The University of Melbourne.
$^{3}$The authors are with the Department of Mechanical Engineering, The University of Melbourne.
        Correspondence: {\tt\small jinghey@student.unimelb.edu.au}}
}

\author{{Jinghe Yang$^{1}$, Mingming Gong$^{2}$, Girish Nair$^{1}$, Jung Hoon Lee$^{3}$, Jason Monty$^{3}$, Ye Pu$^{1}$  }}


\maketitle

\begin{abstract}
In recent years, learning-based feature detection and matching have outperformed manually-designed methods in in-air cases. However, it is challenging to learn the features in the underwater scenario due to the absence of annotated underwater datasets. This paper proposes a cross-modal knowledge distillation framework for training an underwater feature detection and matching network (UFEN). In particular, we use in-air RGBD data to generate synthetic underwater images based on a physical underwater imaging formation model and employ these as the medium to distil knowledge from a teacher model SuperPoint pretrained on in-air images. We embed UFEN into the ORB-SLAM3 framework to replace the ORB feature by introducing an additional binarization layer. To test the effectiveness of our method, we built a new underwater dataset with groundtruth measurements named EASI (https://github.com/Jinghe-mel/UFEN-SLAM), recorded in an indoor water tank for different turbidity levels. The experimental results on the existing dataset and our new dataset demonstrate the effectiveness of our method.

\end{abstract}

\input{1_Intro/introduction}

\input{2_Background/Back_1}

\input{3_Method/method}

\input{4_Results/result}

\bibliographystyle{IEEEtran}
\bibliography{IEEEabrv, ref}
\end{document}

%% file: 1_Intro/introduction.tex
\section{Introduction}
At present, Autonomous Underwater Vehicles (AUVs) are competent for many tasks, such as repairing underwater facilities, natural gas extraction, and environmental safety inspections \cite{intro1}. In order to complete complex control tasks, AUVs need to be able to locate themselves. However, limited sensor choices and poor visualization make localization challenging in the underwater environment. For example, the Global Positioning System (GPS) and high-frequency radio signals are unavailable in underwater environments. Consequently, AUVs must rely on other sensors to obtain their position information, such as cameras and sonars \cite{intro2}. 

Visual SLAM (VSLAM) is one potential vision-based solution to localization and environmental perception in AUVs. VSLAM algorithms use cameras as the main exteroceptive sensor. The front end of VSLAM, named Visual Odometry (VO), accepts the perceived images and finds corresponding features in adjacent frames to estimate the current states \cite{intro5}. VSLAMs can be divided into feature-based and direct methods depending on how the geometric constraints are extracted at the front end \cite{kd5}. Feature-based VSLAM uses sparse feature points to construct 3D landmarks in space, while the direct methods estimate the motion trajectory based on the intensity of pixel values on continuous images. VSLAM is potentially a good candidate for underwater scenarios when AUVs are close to the bottom, and there is enough visibility to observe some objects or details on the seabed. However, the performance of VSLAM is affected by the underwater environment. Joshi et al. \cite{intro6} compared VSLAM algorithms on several underwater sequences. Their results show that the feature-based indirect method ORB-SLAM2 \cite{kd6_2} performs more robust than other methods, e.g., LSD-SLAM \cite{LSD} and DSO \cite{DSO}, but still suffers from the unstable performance of the feature extraction and matching due to the blurriness and low contrast of the underwater images. 

For feature-based VSLAM such as ORB-SLAM2, the quality of the detected features and feature matching between neighbouring frames is a core factor of the performance. In \cite{intro12}, the experimental results show that attenuation and backscattering in underwater imaging reduce the effectiveness of the traditional handcrafted feature detection methods. Recently, learning-based feature extraction and matching have shown great potential for in-air scenarios \cite{kd11} \cite{kd12} \cite{superpoint}. For example, SuperPoint \cite{superpoint} is a self-supervised solution with a good performance on the HPatches dataset \cite{hpatches} compared with the handcrafted methods SIFT \cite{kd10} and ORB \cite{kd7}. However, when applying the learning-based methods to underwater environments, we face new challenges, for example, the absence of suitable underwater datasets for training and the blurriness of the underwater images, making it particularly challenging to define and label the actual feature points.

To tackle these challenges, we leverage the Cross-Modal Distillation \cite{kd3} framework to transfer knowledge from in-air images to underwater images. More specifically, we distil knowledge from a teacher SuperPoint model trained on in-air images to our student underwater feature detection and matching network. To enable efficient knowledge transfer, we propose synthesising underwater images from in-air images using a physical model and designing proper loss functions to train the student network on synthetic underwater images. This framework transfers knowledge from a rich labelled modality as supervision to an unlabeled paired modality. 
Below is the list of our contributions in this paper:
\begin{itemize}
\item We propose the Underwater Feature Extraction Network (UFEN) - an underwater feature point detection and description neural network that directly extracts and matches feature points on underwater images.
\item We use the idea of Cross-Modal Knowledge Distillation in the training of UFEN. We transfer the knowledge of the pre-trained model (SuperPoint) for the in-air scenarios to the UFEN model via synthetic underwater images. 
\item We propose a monocular underwater VSLAM system UFEN-SLAM\footnote{The code of
UFEN-SLAM will be made public shortly.}, in which the UFEN is integrated into the state-of-art feature-based VSLAM framework - ORB-SLAM3 \cite{kd6} by introducing an additional binarization layer to the end of the descriptor decoder.
\item  We built an underwater dataset referred to as EASI with the groundtruth measured by a motion tracking system with different water turbidity levels and evaluated the performance of our method on this dataset.
\end{itemize}

%% file: 2_Background/Back_1.tex
\section{Preliminaries}
This section introduces the three main tools, which will be used in our algorithm, including a state-of-art VSLAM framework - ORBSLAM3 \cite{kd6}, a pre-trained neural network for feature extraction and matching - Superpoint \cite{superpoint}, and an underwater image formation physical model for synthetic data generation.

\subsection{ORB-SLAM3}
A state-of-art feature-based method, ORB-SLAM3 consists of four components, including Atlas for representing the set of disconnected maps and three threads: i) tracking; ii) local mapping; iii) loop and map merging. It uses ORB features \cite{kd7} and the image pyramid to identify spatial feature points at different distances. ORB feature detects corner points by the differences in pixel intensities and uses a 256-dimensional binary descriptor with orientation invariance. ORB is designed for in-air cases. Due to the attenuation of underwater sequences, the sharpness of the image is degraded, and the backscattering caused by particles can give the image different characteristics from in-air images. The experimental results in \cite{intro6} show that on underwater sequences, the monocular ORB-SLAM2 \cite{kd6_2} frequently fails to track because of the unstable performance of the ORB feature. Furthermore, it is hard to initialize if the image is blurry and in low contrast. However, it is still a better candidate than the direct methods, DSO and LSD-SLAM.

In Section \ref{UFEN-SLAM}, we will integrate the newly trained UFEN model into the ORB-SLAM3 framework for detecting feature points and assigning descriptors. 

\subsection{SuperPoint}
\begin{figure}[tbp]
\centerline{\includegraphics[width=0.4\textwidth]{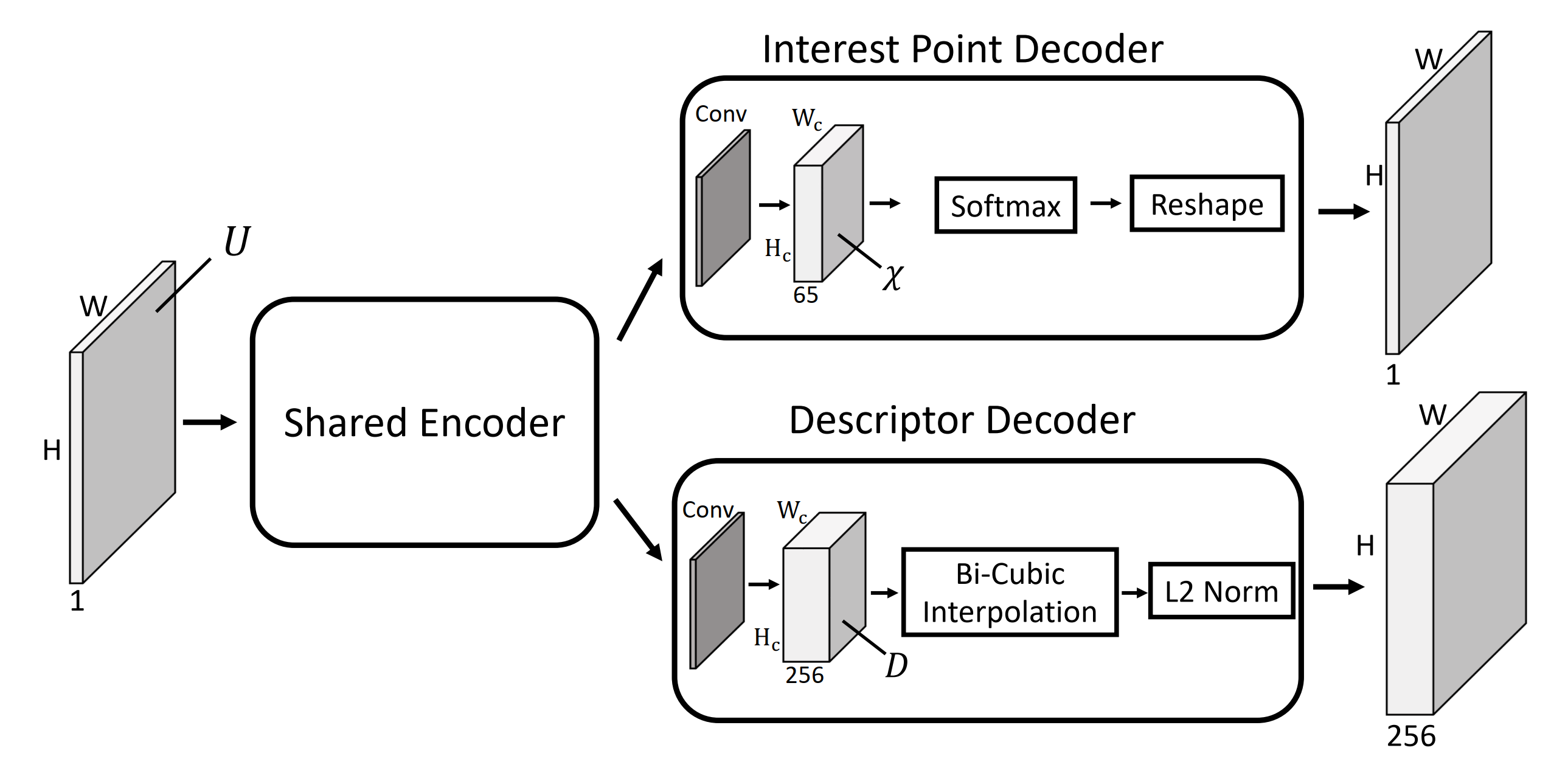}}
\caption{SuperPoint network architecture consists of a shared encoder and two decoders.}
\label{fig:superpoint}
\end{figure}

SuperPoint \cite{superpoint} is a learning-based feature point extracting and matching neural network. As shown in Fig. \ref{fig:superpoint}, SuperPoint has one shared encoder and two decoders, including an interest point detector decoder and a descriptor decoder. SuperPoint takes grayscale images $U \in \mathbb{R} ^{H \times W \times 1}$ of size $H \times W$ as its input. The interest point detector head computes $\mathcal{X} \in \mathbb{R} ^{H_c \times W_c \times 65}$ and outputs a probability tensor $\in \mathbb{R} ^{H \times W \times 1}$ after the channel-wise $Softmax$ and $Reshape$ operations. $H_c$ and $W_c$ are $1/8$ of the input images size. The descriptor decoder head computes $\mathcal{D} \in \mathbb{R} ^{H_c \times W_c \times 256}$ and finally outputs a 256-dimensional descriptor tensor $\in \mathbb{R} ^{H \times W \times 256}$ after the bi-cubic interpolation and L2-normalization.

SuperPoint is initially trained on corner points of the synthetic general shapes. This guarantees the robust performance of SuperPoint when applied to different scenarios, tending to focus on corner points in the environment, i.e. on pixels with significant changes in pixel intensity. We use the pre-trained SuperPoint network in \cite{superpoint} as the teacher model in the knowledge distillation framework to train our UFEN network.

The descriptors given by SuperPoint are not binary, which makes SuperPoint unsuitable for direct integration into ORB-SLAM3 and therefore requires additional binarization operations.

\subsection{Underwater Image Formation}
In water, light reflected from the surface of underwater objects is attenuated through absorption, causing image degradation. In addition, the tiny particles in water can lead to backscattering and veiling appearance in images. An underwater image physical model is proposed in \cite{r6} to simulate these effects:
\begin{equation}
    I_c = J_c \cdot e^{-\beta_c z} + B_c^\infty (1 - e^{-\beta_c z}) .
    \label{equ:1}
\end{equation}

The subscript $c$ represents the three color channels $R$, $G$ and $B$. $z$ is the distance from the camera. $I_c$ and $J_c$ denote the pixel values of the captured underwater image and the clear in-air image, respectively. $B_c^\infty$ is a coefficient for the background light, describing the ambient brightness in water. $\beta_c$ is defined as the beam attenuation coefficient, which depends on the wavelength $\lambda$ of the color channel. The parameter $\beta_c$ in the formation model can be estimated through adaptive learning approaches, such as WaterGAN \cite{li2017watergan}, or by referring to the Jerlov water type \cite{r3}. Jerlov water types classify different underwater scenarios based on the spectral transmittance of downward irradiance. In \cite{r6}, the authors point out that the spectral transmittance in the Jerlov water type is equivalent to the \textit{diffuse attenuation coefficient of the downwelling light} denoted by $K_d(\lambda)$. 

The authors in \cite{r7} show the following relationship between $K_d(\lambda)$ and $B^\infty(\lambda)$, that is:
\begin{equation}
    B^\infty(\lambda) =  \frac{b(\lambda)E(d, \lambda)}{\beta(\lambda)} ,
    \label{background}
\end{equation}
with
\begin{equation}
    E(d, \lambda) = E(0, \lambda) e ^ {-K_d(\lambda)d} ,
    \label{background1}
\end{equation}
 where {$b(\lambda)$ is the beam scattering coefficient. And $E(0, \lambda)$ is the energy of light before the transmission near the water's surface. $E(d, \lambda)$ is the energy after the transmission of a water depth $d$. Henceforward, $\beta(\lambda)$ describes the attenuation of the underwater beam as it transmits horizontally. $K_d(\lambda)$ governs the energy loss of the light beam propagating vertically downward in water, determining the ambient light. These two are different coefficients, but both affect underwater imaging. The values of these two coefficients for different underwater environments can be found in \cite{r3} and \cite{r8}.

%% file: 3_Method/method.tex
\begin{figure}[tb]
\centerline{\includegraphics[width=0.5\textwidth]{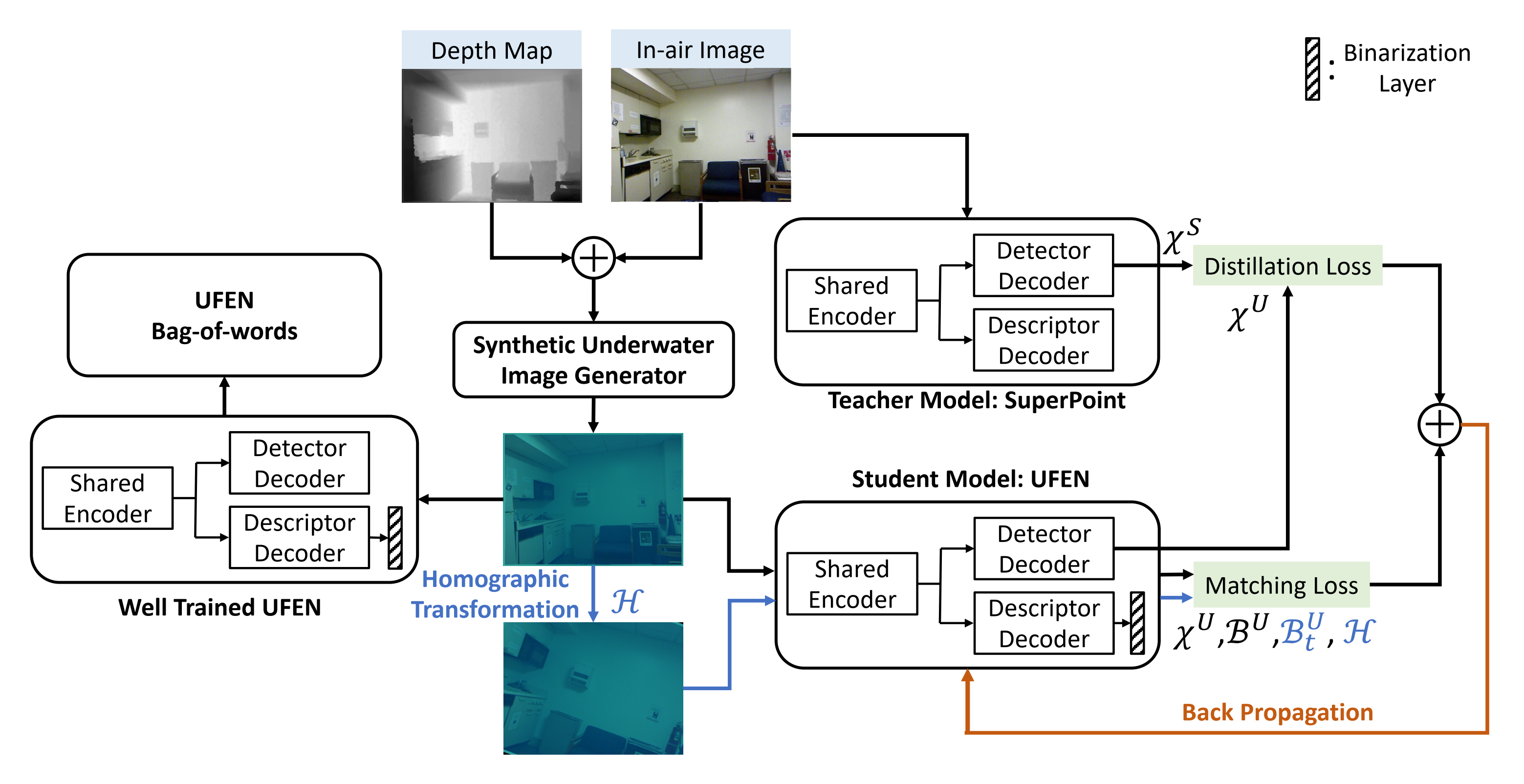}}
\caption{Overall training framework of the UFEN. It needs the paired in-air and depth image as the training inputs. The final well-trained UFEN generates the UFEN Bag-of-words vocabulary based on the synthetic underwater images.}
\label{fig:1}
\end{figure}

\section{Underwater Feature Extraction Network}
We propose the Underwater Feature Extraction Network (\textbf{UFEN}) to reduce the effect of backscattering and attenuation in extracting and matching features on underwater images. Later in Section \ref{UFEN-SLAM}, we integrate the UFEN model into ORB-SLAM3. Fig. \ref{fig:1} shows the overall offline training framework. The first component is the synthetic underwater image generator based on a physical model using depth and in-air images (see Section \ref{imaging}). To successfully integrate the trained UFEN model into the ORB-SLAM3 framework, we introduce an additional binary layer to the UFEN model. In the training of UFEN, we used the joint training of cross-modal knowledge distillation and correspondence matching to guarantee the feature extraction capability and the binary descriptor matching (see Section \ref{network}). 
\subsection{Synthetic Underwater Image Generator} \label{imaging}
\begin{figure}[tbp]
\centerline{\includegraphics[width=0.5\textwidth]{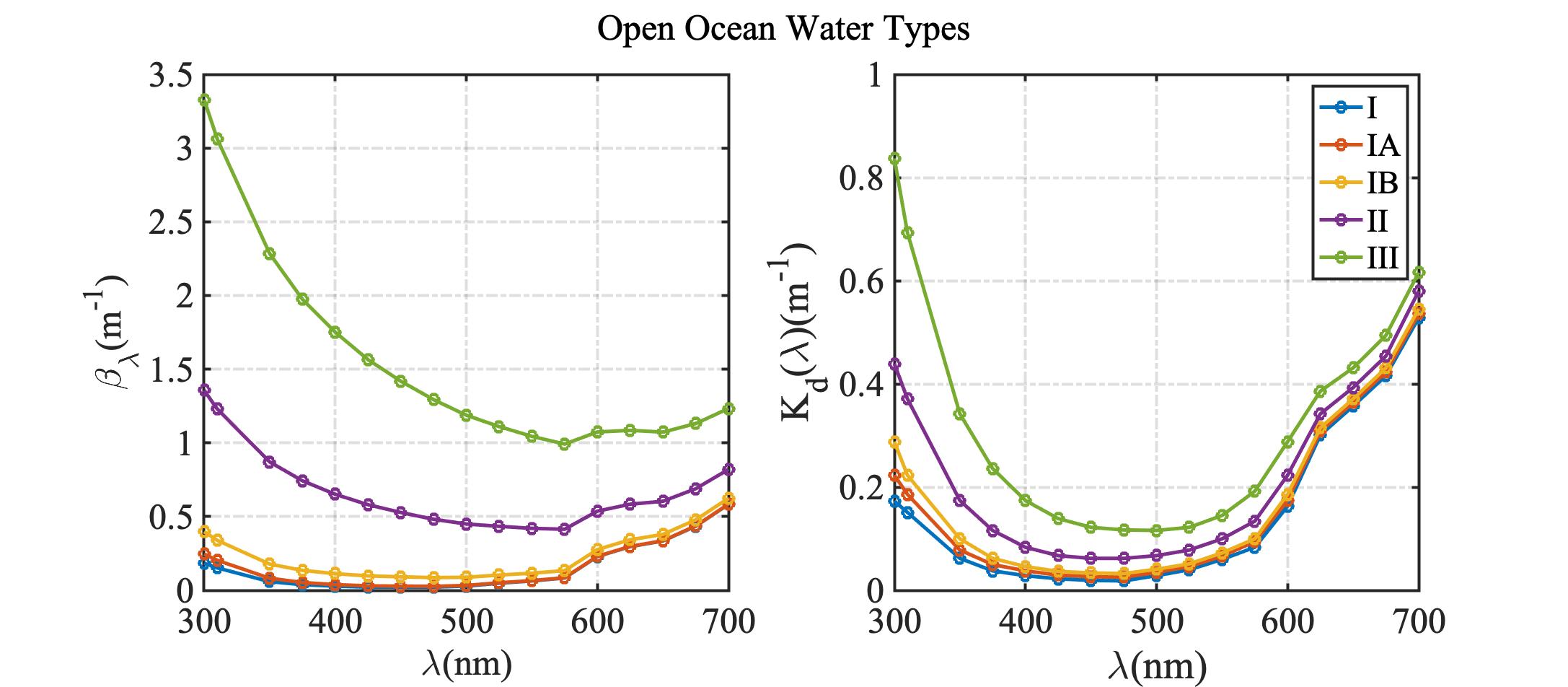}}
\caption{The parameter values $\beta_{\lambda}$ and $K_d(\lambda)$ in Open Ocean Water Types from Type I to Type III.}
\label{fig:para}
\end{figure}

We propose to combine the physical models in \eqref{equ:1}, \eqref{background} and \eqref{background1} to set a connection of the background light with the water characteristics. Then we have,
\begin{equation}
    I_c = J_c e^{-\beta_c z} + \frac{b_c E(0, \lambda_c) e ^ {-K_d(\lambda_c)d}} {\beta_c}(1 - e^{-\beta_c z}) + W,
    \label{physics_model}
\end{equation}
where $I_c$ is the pixel value of the generated synthetic underwater image, determined by the wavelength $\lambda_c$ of the color channel $c$. $W$ is a random Gaussian noise over the image plane that simulates the flow of particles in water. The background light $B_c^\infty$ in \eqref{equ:1} is estimated via \eqref{background} and \eqref{background1} with the coefficients $K_d(\lambda_c)$, $\beta_c$ and $b_c$. We use the parameters suggested in \cite{r3} and \cite{r8} to generate synthetic underwater images close to the natural underwater scenarios. The value ranges of the parameters $\beta_{\lambda}$ and $K_d$ of the five open ocean water types are shown in Fig.~\ref{fig:para}. We utilized water types $\uppercase\expandafter{\romannumeral1}$ and $\uppercase\expandafter{\romannumeral3}$ as the lower and upper bounds, and take random samples of $\beta_\lambda$ and $K_d$ from these intervals and apply them in \eqref{physics_model} to generate synthetic images, to encompass the various underwater environments.
\subsection{Underwater Feature Extraction Network (UFEN)} \label{network}
\subsubsection{Network Structure}
UFEN is based on the same structure as SuperPoint, with an additional binarization layer at the end of the descriptor decoder. The network input is the grayscale image $U \in \mathbb{R} ^{H \times W \times 1}$. The heads of the detector and descriptor decoders compute $\mathcal{X} \in \mathbb{R} ^{H_c \times W_c \times 65}$ and  $\mathcal{D} \in \mathbb{R} ^{H_c \times W_c \times 256}$, respectively, and finally output the tensors with the same size as the input gray image. The binarized descriptor tensor is denoted by $\mathcal{B} \in  	\left\{ 0,1  \right\} ^{H \times W \times 256}$.

\subsubsection{Binary Activation Layer}
An additional binary activation layer is embedded at the end of the descriptor decoder of UFEN to fit the binary descriptor format in ORB-SLAM3. Inspired by GCNv2 \cite{kd9}, we use the method described in \cite{m1} to propagate the gradients through the straight-through estimator to avoid backpropagating the constant zero gradients. The binary activation function $\textbf{b}(x)$ and the straight-through estimator $g_b$ of the gradient are as follows:
\begin{equation}
    \begin{aligned}
    & \text{Forward:  } \textbf{b}(x) = Sign(x) =
        \begin{cases}
        + 1 &\text{if x $\geq$ 0,} \\
        - 1 &\text{Otherwise,}
        \end{cases}\\ 
    &\text{Backward:  } g_b = 1_{|x|\leq1}  .      
    \end{aligned}
    \label{binary}
\end{equation}

We apply the activation function in \eqref{binary} to every element and get $\mathcal{B}$. Since the descriptor vector is L2-normalized before the binarization, it is in the range $(-1, +1)$ throughout.

\subsubsection{Loss Functions}
The loss functions aim to perform cross-modal knowledge distillation from SuperPoint to UFEN, in order to enable UFEN to achieve comparable performance on underwater images as SuperPoint does on in-air images, minimising the impact of low visibility and turbidity in water on feature detection.

The loss function used in this paper consists of two parts $L_{p}$ and $L_{d}$. The superscript $S$ and $U$ are used to denote the outputs given by SuperPoint and UFEN, respectively. The first component $L_{p}$ is for the knowledge distillation between the computed tensor of the detector decoder of SuperPoint $\mathcal{X}^S$ and that of UFEN $\mathcal{X}^U$. The knowledge of binary descriptors cannot be directly distilled from SuperPoint. Therefore we adopt the homographic transformation $\mathcal{H}$ to simulate the change of the viewpoint and use $L_{d}$ to guarantee the binary descriptor matching performance through the matching loss of corresponding feature points between images. $\mathcal{B}^U$ and $\mathcal{B}^U_{t}$ are the outputs of the UFEN descriptor decoder on the original image and the transformed images. We use the coefficient  $\alpha$ to adjust the training weights of the two components:
\begin{multline}
L(\mathcal{X}^S, \mathcal{X}^U, \mathcal{B}^U, \mathcal{B}^U_{t}, \mathcal{H}) = \\
 L_{p}(\mathcal{X}^S, \mathcal{X}^U) + 
   \alpha L_d(\mathcal{X}^U, \mathcal{B}^U, \mathcal{B}^U_{t}, \mathcal{H}).
   \label{final_loss}
\end{multline}

The loss component $L_{p}$ tends to make SuperPoint and UFEN's detector decoder output the close probability maps and select similar feature points. To achieve this goal, we propose to use the Kullback–Leibler (KL) divergence to distil a "soft target" and adopt the Probabilistic Knowledge Transfer (PKT) \cite{m2} to distil the global feature distribution. In particular, the loss function $L_{p}$ is set to be:
\begin{equation}
    L_{p}(\mathcal{X}^S, \mathcal{X}^U) = L_{KL}(\mathcal{X}^S, \mathcal{X}^U) + \beta L_{PKT}(\mathcal{X}^S, \mathcal{X}^U),
\end{equation}
with
\begin{equation}
    L_{KL}(\mathcal{X}^S, \mathcal{X}^U) = \frac{1}{H_c W_c}
    \sum_{\substack{h=1 \\ w=1}}^{H_c, W_c} D_{KL}(\sigma(x^S_{hw}), \sigma(x^U_{hw})),
\end{equation}
in which $x^S_{hw}$ $\in \mathbb{R} ^{65}$ and $x^U_{hw}$ $\in \mathbb{R} ^{65}$ are the vectors in $\mathcal{X}^S$ and $\mathcal{X}^U$, and $\sigma(\cdot)$ is the $Softmax$ function that converts a real number vector to a probability distribution. The function $D_{KL}$ is defined as,

\begin{equation}
    D_{KL}(s,u) = \sum_{k=1}^{65} s_k \cdot (log(s_k) -log(u_k)),
\end{equation}
with $s=[s_1,s_2,\dots,s_{65}]$ and $u=[u_1,u_2,\dots,u_{65}]$ ($s \geq 0$, $u\geq0$, $\sum s_k = 1$, $\sum u_k = 1$). The KL loss quantifies the difference between two probability maps element-by-element but lacks the correlation information between the elements. Significantly, the image blur on underwater images distributes unevenly, where near places are clear and more distant places are blurred. Therefore, we use PKT loss \cite{m2} to help UFEN to have a high-level view of the feature point detection knowledge on an underwater image and learn the global information. The PKT loss distils the probabilistic knowledge from a teacher model to a student model by minimizing the divergence of the probability distribution given by the output of the two models. It uses the cosine distances between the feature vectors to reflect the distribution pattern. The PKT loss is defined as follows:
\begin{equation}
     L_{PKT}(\mathcal{X}^S, \mathcal{X}^U) = \sum_{\substack{i=1}}^{H_c W_c} \sum_{\substack{j=1, i\neq j}}^{H_c  W_c} s_{j|i}\log(\frac{s_{j|i}}{u_{j|i}}),
\end{equation}
where $s_{j|i}$ and ${u_{j|i}}$ are the conditional probability distribution of $\mathcal{X}^S$ and $\mathcal{X}^U$. It is defined as:
\begin{equation}
    p_{i|j} = \frac{K(x_i, x_j)}{\sum^{H_c W_c}_{k=1, k \neq j} K(x_k, x_j)} \in [0, 1],
\end{equation}
where $x$ is the L2-normalized feature vector $\in \mathbb{R}^{65}$ extracted from $\mathcal{X}^S$ or $\mathcal{X}^U$ after $Softmax$ and $K(x_i, x_j)$ is the similarity metric and calculated by the cosine distance between vectors,
\begin{equation}
    K(a, b) = \frac{1}{2} (\frac{a^Tb}{||a||_2||b||_2} + 1) \in [0, 1].
\end{equation}

We use the matching loss $L_{d}$ to optimize the matching performance of the binary descriptors. The original image and the transformed image are the inputs of the network UFEN as shown in Fig. \ref{fig:1}. We first find the feature points on the original image using $\mathcal{X}^U$, use $\mathcal{H}$ to project these points to the transformed image, and then find all matching points between the two images. We notate the matching points on the original image and transformed image as $x_i$ and $x_i^t$ for $i= 1\dots N$, respectively. We denote the descriptors corresponding to $x_i$ and $x_i^t$ as $d_i$ and $d_i^t$, respectively. $N$ is the number of matching points. The superscript $t$ is used for the points and descriptors on the transformed image. 

We further define the non-matching points of each $x_i$. Given a $x_i$ on the original image,  a point $x^t_j$ on the transformed image is a non-matching point of $x_i$ if $j\neq i$ and $|x^t_j- x_i^t|>T$, where $T$ is a pre-specified constant threshold. We use $x^t_{i,j}$ to denote the non-matching points of $x_i$. Correspondingly, the descriptor of $x^t_{i,j}$ is denoted by $d^t_{i, j}$. Note that all $d_i$, $d_i^t$, $d^t_{i,j}$ are binary vectors with dimension equal to $1\times 256$ extracted from $\mathcal{B}^U$ or $\mathcal{B}^U_{t}$.

The loss $L_{d}$ is similar to the loss function proposed in \cite{m3} to minimise the descriptor distance between $d_i$ and $d^t_i$ for all matching points, i.e., $i=1\dots N$ and maximise the descriptor distance between $d_i$ and the descriptors of the non-matching points $d^t_j$. $L_{d}$ is defined as: 
\begin{equation}
 L_{d}(\mathcal{X}^U, \mathcal{B}^U, \mathcal{B}^U_{t}, \mathcal{H}) = \frac{1}{N}\sum^N_{i=1}(p_i^2 + n_i^2),
\end{equation}
where
\begin{equation}
\begin{aligned}
& p_i = \max(0, \text{dist}(d_i, d^t_i) - P),\\
& n_i = \max (0, Q - \min(\text{dist}_n(d_i), \text{dist}_n(d^t_i))),
\end{aligned}
\end{equation}
The constants $P$ and $Q$ are two distance margins to the acceptable descriptor distance of matching points and non-matching points. The function $\text{dist}(\cdot)$ is the hamming distance - the number of different symbols between two binary descriptors. $\text{dist}_n(d_i)$ finds the minimal descriptor distance between $d_i$ to the descriptors of the non-matching point set $d^t_{i,j}$,

\begin{equation}
    \text{dist}_n(d_i) = \min( \text{dist}(d_i, d^t_{i,j})). 
\end{equation}

\section{UFEN-SLAM} \label{UFEN-SLAM}
\begin{figure}[tbp]
\centerline{\includegraphics[width=0.4\textwidth]{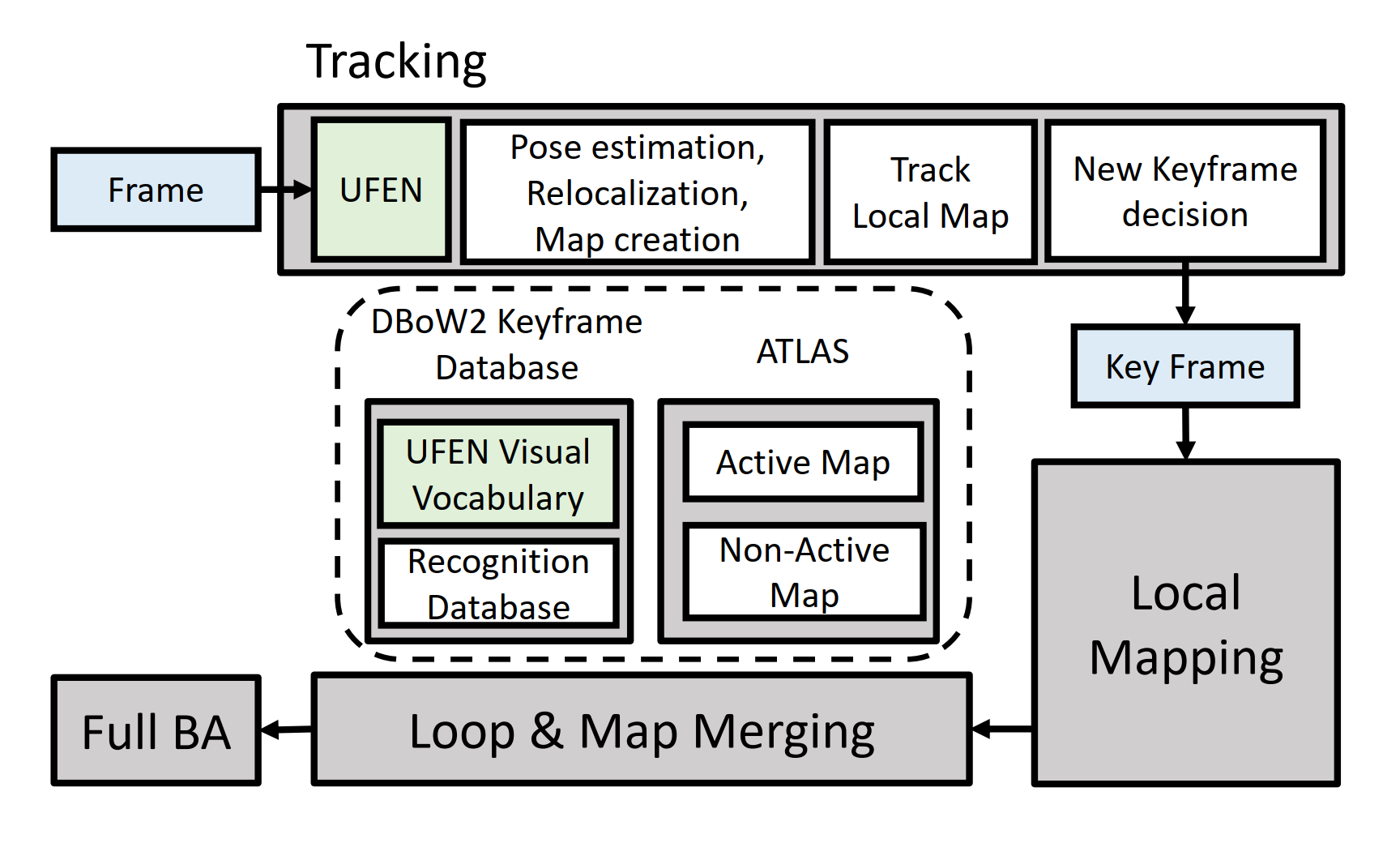}}
\caption{UFEN-SLAM system framework}
\label{fig:2}
\end{figure}

The framework of UFEN-SLAM is shown in Fig. \ref{fig:2}. UFEN-SLAM is based on the monocular ORB-SLAM3 framework, where the trained UFEN replaces the ORB feature in ORB-SLAM3 (green blocks). UFEN has two decoders. The detector decoder extracts feature points while the other decoder assigns a 256-dimensional binary descriptor to each feature point. After detecting feature points and finding the correspondences, we follow the rest steps in ORB-SLAM3. Referring to the integration method in GCNv2 \cite{kd9}, we also use nearest-neighbour search matching in the reference keyframe tracking, replacing the bag-of-words matching in ORB-SLAM3, to show the actual feature extraction and matching performance of UFEN. Finally, UFEN regenerates the bag-of-words vocabulary on synthetic underwater images for loop closure detection.

%% file: 4_Results/result.tex
\begin{figure}[bp]
\centerline{\includegraphics[width=0.4\textwidth]{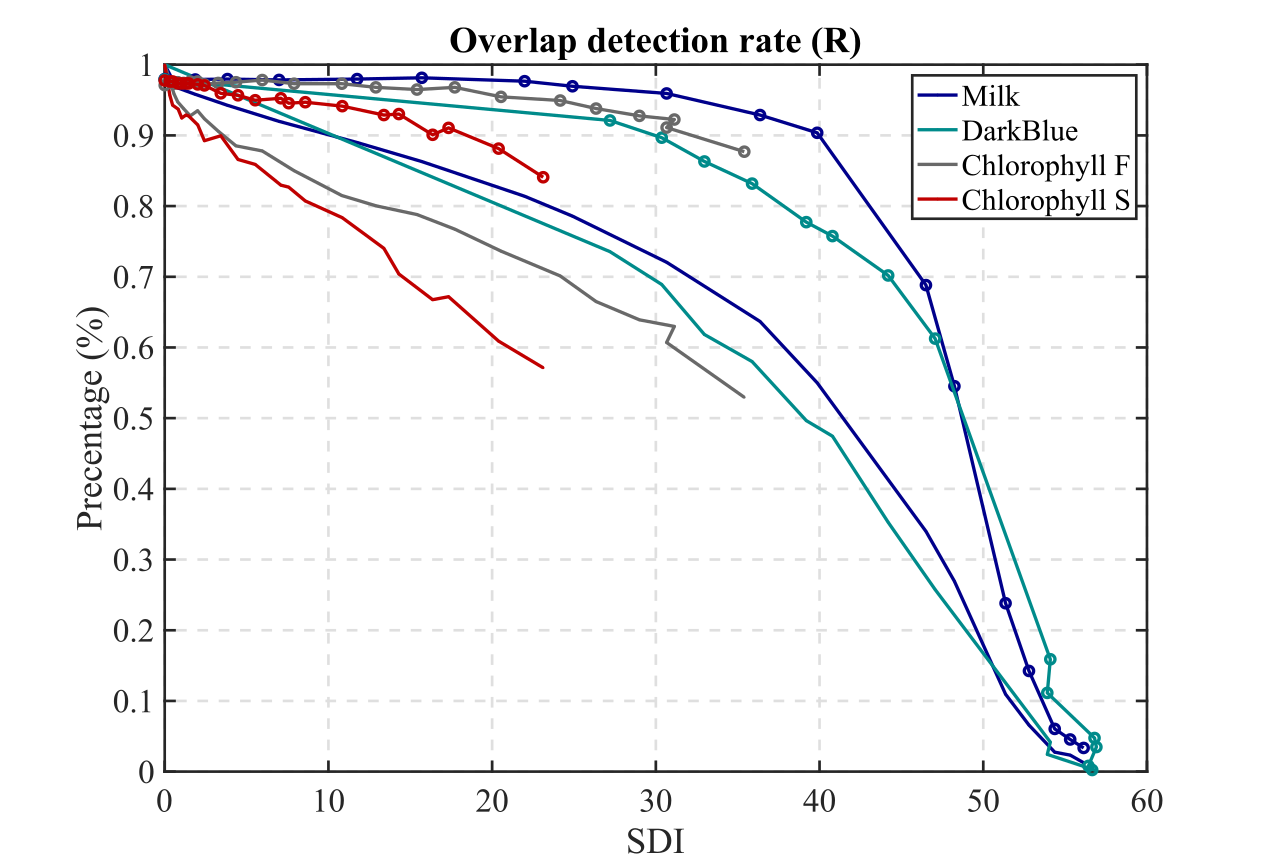}}
\caption{Feature detection evaluation on TURBID dataset. The lines with circle markers are the results of UFEN, others are the results of SuperPoint.}
\label{fig:3}
\end{figure}

\section{EASI Underwater Dataset} \label{EASIDATA}
To investigate the influence of water turbidity on the performance of underwater VSLAM, we collected EASI underwater image sequences with groundtruth measured by the Vicon tracking system in the Extreme Air-Sea Interaction (EASI) Facility in Michell Hydrodynamics Laboratory at the University of Melbourne.

\textbf{Experimental Setup: } The EASI facility is a water tank with 1.8$m$ width and 1.1$m$ water depth. We installed four Vicon camera tracking systems (Vicon Vero, 2.2MP) above the tank, giving it an overlapping field-of-view area of nearly 3$m$ length. Some green artificial plants and marine life images were fixed on a solid base plate at the bottom of the tank. We designed a
rigid frame attached with a GoPro camera on one side and spatially distributed tracking markers on the other side. During the acquisition process, the camera moves in water, and the tracking markers stay above the water and are tracked by the Vicon system.

\begin{figure}[t]
\centerline{\includegraphics[width=0.5\textwidth]{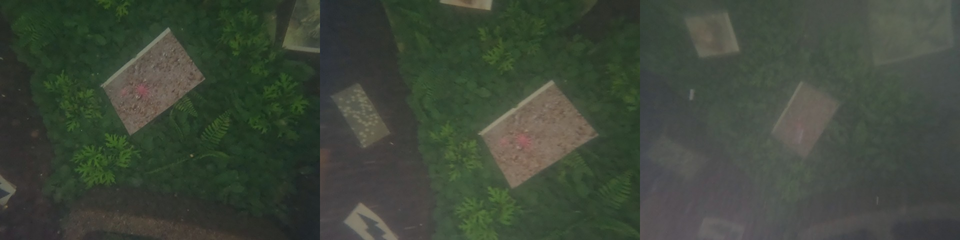}}
\caption{Image samples in EASI dataset of three turbidity levels. From left to right: Clear, Medium, Heavy.}
\label{fig:EASI}
\end{figure}

\textbf{Data Collection:}
The data sequences were initially collected with a GoPro Hero 10 with a sampling rate of 30 Hz and 4K resolution and downsampled to $640 \times 480$. GoPro HyperSmooth mode was disabled when acquiring the raw videos. The Vicon system measures the ground truth with a 100 Hz acquisition rate. The dataset consists of ten sequences with three different turbidity levels ranging from 0 to 5 NTU (Nephelometric Turbidity Units). Image samples are shown in Fig. \ref{fig:EASI}. We poured different amounts of talcum powder into water to control the turbidity, quantified using a YSI ProDSS turbidity meter. 

\begin{table*}[ht]
\caption{ATE(m) of VSLAMs on EASI Dataset}
\begin{center}
\begin{tabular}{|c|c|c|c|c|c|c|c|}
\hline
\textbf{Data}&\textbf{Data}& \textbf{LED}& \textbf{TUR}& \textbf{\textit{ORB}}&  \textbf{\textit{SuperPoint}} & \textbf{\textit{UFEN (w/o CMD)}} & \textbf{\textit{UFEN}}\\
\textbf{Seq.} & \textbf{Length} & \textbf{Light}& \textbf{(NTU)}  & \textbf{\textit{SLAM3}}& \textbf{\textit{SLAM}} & \textbf{\textit{SLAM}} & \textbf{\textit{SLAM}} \\
\hline
01 & 1'\,14'' & On & Clear & 0.0198 & 0.0202 & \textbf{0.0191} & 0.0196\\ 
02 & 0'\,59'' & On & (0-2) & \textbf{0.0181} & 0.0195 & 0.0198 & 0.0205 \\ 

\hline

03 & 1'\,18'' & On & \multirow{3}{*}{\begin{tabular}[c]{@{}c@{}}Medium \\ (2-3.5)\end{tabular}}  &  0.0197 & 0.0171 & 0.0174 & \textbf{0.0168}\\ 
04 & 1'\,32'' & On & &  0.0178 & 0.0207 & 0.0177 & \textbf{0.0152}\\ 
05 & 1'\,24'' & Off & & F & F & 0.0236 & \textbf{0.0203} \\ 

\hline

06 & 1'\,14'' & On & \multirow{5}{*}{\begin{tabular}[c]{@{}c@{}}Heavy\\ (3.5-5)\end{tabular}} 
& 0.0192 & 0.0226 & 0.0198 & \textbf{0.0173}\\ 
07 & 1'\,22'' & On & & 0.0241  & 0.0255 & 0.0235& \textbf{0.0234}\\ 
08 & 1'\,11'' & On & & 0.0187  & 0.0185 & 0.0173& \textbf{0.0166}\\ 
09 & 1'\,20'' & On & &  F & F & F& \textbf{0.0152} \\ 
10 & 1'\,27'' & On & & F  & F & F& \textbf{0.0193}\\ 
\hline
\end{tabular}
\label{tab3}
\end{center}
\end{table*}

\section{Experimental Results}

In this section, we present our experimental results from two perspectives to demonstrate the effectiveness of our approach. These are the evaluation of the feature points detection using the TURBID Dataset \cite{e0}, and an experimental comparison based on the EASI dataset introduced in section \ref{EASIDATA}. 

\subsection{Training Datasets}
The training data of UFEN is based on NYU depth V2 dateset\cite{NYU} and DIML/CVL RGB-D indoor Dataset\cite{DIML}. The depth maps in DIML/CVL indoor dataset are filled using a colour-guided depth upsampling algorithm proposed in \cite{DIML1}. All the images are forcibly downsampled to $640\times480$.

We use the physical model in \eqref{physics_model} when generating synthetic underwater images. For avoiding the dissipation of all image information, the maximum image depth is set to be $3m$. Meanwhile the depth of the water $d$ is assumed to be $5m$ and the water surface ambient light $E(0, \lambda)$ is assumed to be perfectly white. The wavelength of the colour channels $B$, $G$, $R$ is chosen to be $450 nm$, $525 nm$ and $700 nm$ respectively.

\subsection{Feature Detection Evaluation on TURBID dataset} \label{Turibidity}
This evaluation is based on an underwater colour restoration dataset - the TURBID Dataset \cite{e0}, which mixes different amounts of milk, blue ink and chlorophyll to adjust the turbidity level. We exemplify the model performance of UFEN and SuperPoint by comparing their overlap detection rate of the reference feature points at different turbidity levels. The detection threshold for both models has been set to 0.01. We define the reference feature points as those detected by SuperPoint on the clear image. The overlap detection is the detected point on the turbid images, which is in a $3$ by $3$ pixel grid centred on the reference point. We use overlap detection rate $R$ as the evaluation metric. That is, $R = \frac{P_c}{P_R}$. $P_{R}$ is the number of the reference feature points, and $P_c$ is the number of the overlap detection. We adopt SDI (Structural Degradation Index) \cite{e6} to quantify the turbidity level of the images. The maximum SDI of Milk and DeepBlue is up to nearly 60, until all pattern information disappears. The SDI of chlorophyll datasets provided by TURBID is from 0 to 40. Fig. \ref{fig:3} visualises the varying detection performance across the different turbidity levels.

Our model UFEN has a larger $R$ than the teacher model SuperPoint in all cases. In particular, When the SDI is less than 40, the decrease of $R$ of UFEN is much smoother than SuperPoint. When the image is too fuzzy (SDI $\geq$50), such as at the end of the "Milk" and "DarkBlue" data, both UFEN and SuperPoint cannot detect enough feature points and $R$ approaches zero. The result shows that the performance of the feature detection of UFEN is less affected by the turbidity compared with SuperPoint.

\subsection{VSLAM Evaluation on EASI Dataset}
In this section, we test the proposed UFEN-SLAM using the EASI dataset, and compare it with ORBSLAM3 and SuperPoint-SLAM. In SuperPoint-SLAM, SuperPoint with an additional binarization layer is retrained using the loss function in \eqref{final_loss}, but without synthetic images.

To solve the issue of the scale uncertainty in the monocular VSLAM, we use Umeyama's method \cite{e3} to align the estimated trajectories and find the similarity transformation $S'=\left\{ s', R', t'\right\}$, where $R'$, $s'$ and $t'$ are the rotational matrix, estimated scale, and translation matrix, respectively. Then the quality of the estimated trajectory is evaluated by the Absolute Trajectory Error (ATE) \cite{e4}.
\begin{equation}
    ATE_{pos} = (\frac{1}{N} \sum^{N-1}_{i=0}||p_i - (s'R'\hat{p_i}+t')||^2)^\frac{1}{2}
    \label{ATE}
\end{equation}
where $\left\{ p_i\right\}^{N-1}_{i=0}$ is the ground-truth position sequences and $\left\{\hat{p_i}\right\}^{N-1}_{i=0}$ is the estimated positions given by VSLAM algorithms. GoPro and Vicon are two separate systems, so to solve the time synchronization issue, we tune the time interval distance between two systems to find the minimal ATE. 

We manually tune the thresholds in the ORB feature, to achieve the best performance of ORB-SLAM3 in turbid water. The experimental results are shown in Table \ref{tab3}. It shows that UFEN-SLAM performs more robust than ORB-SLAM3 and SuperPoint-SLAM by achieving lower ATEs and more successful tracking on the data sequences at the Heavy and Medium turbidity levels. ORB-SLAM3 fails to initialize on the sequence Heavy-10 and Medium-05, and fails to finish the tracking of Heavy-09. We note that Medium-05 is recorded with a darker light condition, which makes the tracking more challenging. When the water body is clear, ORB-SLAM3 and SuperPoint-SLAM show comparable performance with UFEN-SLAM. Meanwhile, to assess the efficacy of knowledge distillation, we retrained the student model without distilling the cross-modal knowledge of SuperPoint on the in-air images. Instead, we directly fed the synthetic images to SuperPoint and aimed to tune the matching performance of UFEN on the synthetic underwater images without losing performance in detection. Our results indicated that UFEN without Cross-Modal Distillation (w/o CMD) resulted in moderate improvements compared to SuperPoint-SLAM on Medium and Heavy levels. It has lower ATEs on Heavy sequences than SuperPoint-SLAM, and unlike SuperPoint-SLAM and ORB-SLAM3, UFEN (w/o CMD) SLAM can track the Medium-05 sequence successfully. However, UFEN (w/o CMD) SLAM again fails to track Heavy-09 and Heavy-10 and has higher ATEs than UFEN-SLAM on both Heavy and Medium-level sequences. These results demonstrate that UFEN without Cross-Modal Distillation can still slightly improve underwater tracking performance compared to the teacher model. To summarise,  Cross-Modal Distillation can make the model more robust in a turbid underwater environment. That proves the effectiveness of  Cross-Modal Distillation in our training.

\section{Conclusion}
This paper proposes a monocular UFEN-VSLAM for turbid underwater scenes. Moreover, we collect the EASI underwater dataset with multiple turbidity levels. Our approach performs effectively on TURBID and EASI datasets. 


\section*{Acknowledgments}
Authors thank Mr. Max Rounds for the technical support in the experiments. Authors gratefully acknowledge funding support from Australian government grants AUSMURIB000001, DE220101527 and DE210101624.